\documentclass[runningheads]{llncs}

\usepackage{graphicx}
\usepackage{booktabs}
\usepackage[dvipsnames]{xcolor}
\usepackage{multirow}
\usepackage{pifont}
\usepackage{adjustbox}
\usepackage{amsmath}
\usepackage{amssymb}
\usepackage{bbm}
\usepackage{subfig}
\usepackage{array}
\usepackage{wrapfig}
\usepackage{graphicx}
\usepackage{tabularray}
\UseTblrLibrary{booktabs}

\newcommand{\cmark}{\ding{51}}
\newcommand{\xmark}{\ding{55}}
\newcommand{\colorred}{\color{Red}}
\newcommand{\colorgreen}{\color{ForestGreen}}
\newcommand{\colorgrey}{\color{darkgray}}
\newcommand{\colorblack}{\color{Black}}

\usepackage[sortcites,url=false,isbn=false,doi=false,eprint=false,backend=biber,style=lncs]{biblatex}
\bibliography{bibliography}

\usepackage{hyperref}

\newcommand{\xzeropred}{\emph{$x_0$-prediction}\;}
\newcommand{\coneprojection}{\emph{cone-projection}\;}
\newcommand{\req}[1]{\textbf{#1}} 

\begin{document}

\title{Diffusion-based Visual Counterfactual Explanations - Towards Systematic Quantitative Evaluation
\thanks{Supported by the Bavarian HighTech agenda and the
Würzburg Center for Artificial Intelligence and Robotics (CAIRO).}}

\titlerunning{Diffusion-based VCEs - Systematic Quantitative Evaluation}

\author{Philipp~Väth\inst{1,2}\orcidID{0000-0002-8247-7907}
\and
Alexander~M.~Frühwald\inst{1}\orcidID{0009-0007-5347-6239}
\and
Benjamin~Paassen \inst{2}\orcidID{0000-0002-3899-2450}
\and
Magda~Gregorova\inst{1}\orcidID{0000-0002-1285-8130}}

\authorrunning{Vaeth et al.}
\institute{Center for Artificial Intelligence and Robotics, Technical University of Applied Sciences Würzburg-Schweinfurt, Franz-Horn-Straße 2,
Würzburg, Germany\\
\email{\{philipp.vaeth,alexander.fruehwald,magda.gregorova\}@thws.de} \and 
Bielefeld University, Universitätsstraße 25, Bielefeld, Germany\\
\email{\{bpaassen\}@techfak.uni-bielefeld.de}}

\maketitle 
\begin{abstract}
Latest methods for visual counterfactual explanations (VCE) harness the power of deep generative models to synthesize new examples of high-dimensional images of impressive quality.
However, it is currently difficult to compare the performance of these VCE methods as the evaluation procedures largely vary and often boil down to visual inspection of individual examples and small scale user studies.
In this work, we propose a framework for systematic, quantitative evaluation of the VCE methods and a minimal set of metrics to be used.
We use this framework to explore the effects of certain crucial design choices in the latest diffusion-based generative models for VCEs of natural image classification (ImageNet).
We conduct a battery of ablation-like experiments, generating thousands of VCEs for a suite of classifiers of various complexity, accuracy and robustness.
Our findings suggest multiple directions for future advancements and improvements of VCE methods.
By sharing our methodology and our approach to tackle the computational challenges of such a study on a limited hardware setup (including the complete code base), we offer a valuable guidance for researchers in the field fostering consistency and transparency in the assessment of counterfactual explanations.
\keywords{Explainability \and XAI \and Visual Counterfactual Explanations \and VCE \and Generative Modeling \and Diffusion}
\end{abstract}

\section{Introduction}\label{sec:Intro}
One of the greatest challenges of modern machine learning, especially deep neural networks, is maintaining the trust of the public and application experts in the results and recommendations of these powerful technologies \cite{trustworthy_ai}.
This is essential in sensitive and high-risk application domains such as medicine \cite{ai_healthcare_1,ai_healthcare_2}, law enforcement \cite{ai_lawenforcement_1}, and finance \cite{ai_finance}. 
To increase confidence in the reliability of these models and to enhance their transparency, a new research field called Explainable AI (XAI) has been established in recent years \cite{xai_survey}.

The goal of early XAI approaches has been to explain the internal workings of models, such as the gradient backpropagation dynamics \cite{lrp,deconv,grad_cam}. 
Later the focus has shifted towards the explanation of model outputs rather than that of the internal structures of the model \cite{lime, DBLP:conf/pkdd/CasalicchioMB18,DBLP:conf/aaai/Ribeiro0G18,vce}.
For example, multiple popular explanation methods highlight features of the input data that contribute the most to the current prediction \cite{lime,lrp}.
By contrast, counterfactual explanations highlight necessary changes in the input to achieve a different output \cite{ce}. We focus in particular on counterfactual explanations for image data (visual counterfactual explanations; VCE) \cite{vce}.

To create such VCEs, modern deep generative models are particularly promising as they are designed to synthesize new high-dimensional examples by sampling.
The best quality in generative models to date has been achieved by the so-called diffusion models.
Quite naturally, it has recently been also proposed to harness the power of the diffusion models for generating counterfactuals \cite{dvce,dime}.

\begin{figure}[h]
    \centering
    \begin{adjustbox}{max width=\textwidth}
    \begin{tabular}{cccc}
    \toprule
     & \shortstack{Cheetah $\rightarrow$ Leopard} & \shortstack{Mashed Potato $\rightarrow$ Carbonara} & \shortstack{Burrito $\rightarrow$ Pizza} \\
    \midrule
    \shortstack{Counterfactual\\Example\\\vspace{0.3cm}} & \includegraphics[trim={1cm 11cm 1cm 11cm},clip,width=0.3\textwidth]{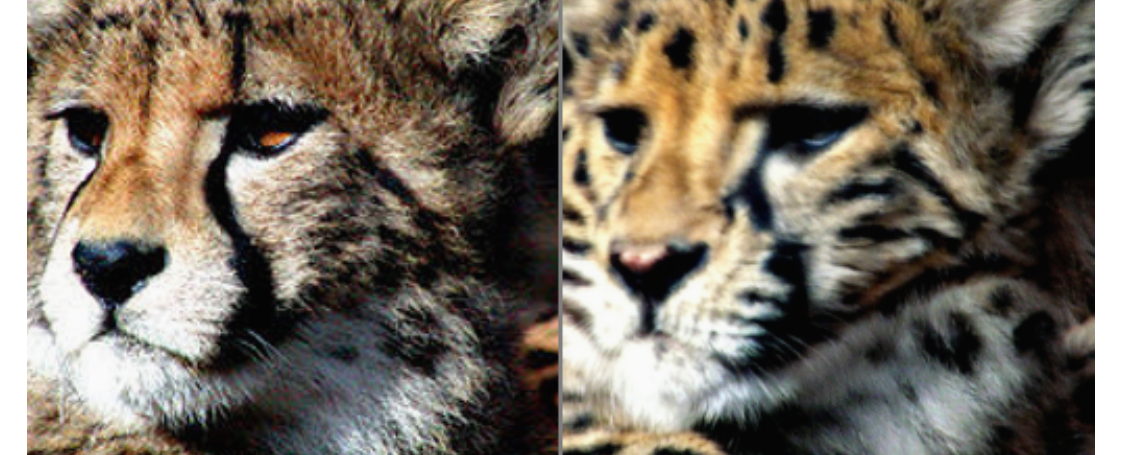} & \includegraphics[trim={1cm 11cm 1cm 11cm},clip,width=0.3\textwidth]{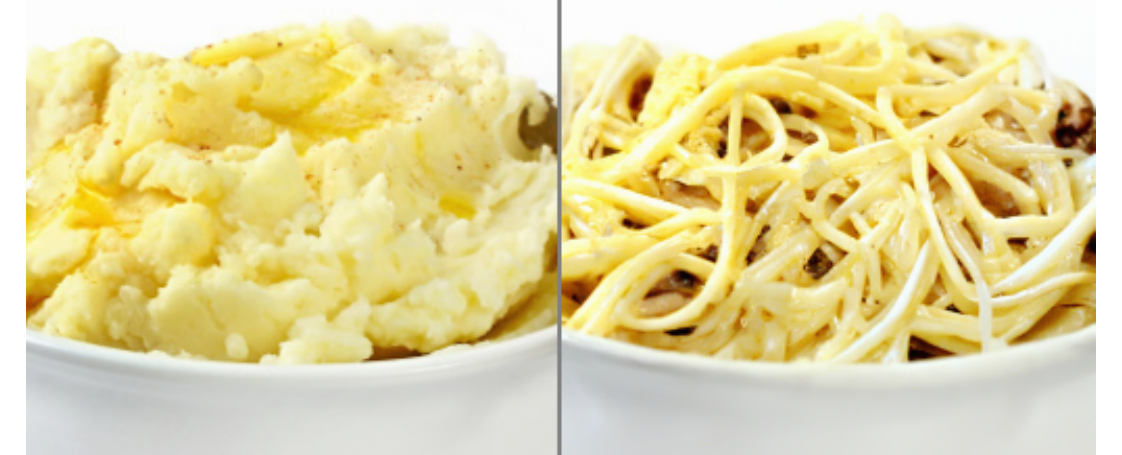} & \includegraphics[trim={1cm 11cm 1cm 11cm},clip,width=0.3\textwidth]{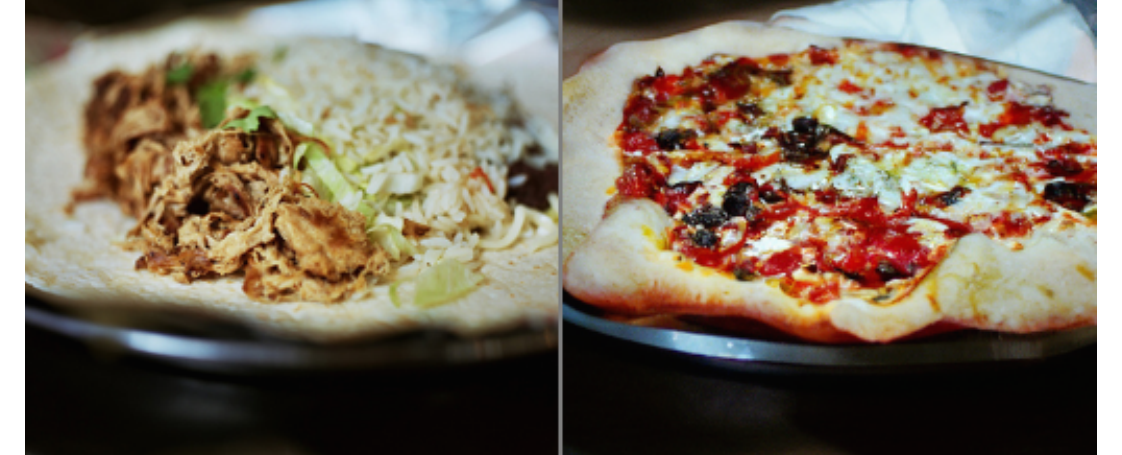} \\
    \shortstack{Adversarial\\Example\\\vspace{0.3cm}} & \includegraphics[trim={1cm 11cm 1cm 11cm},clip,width=0.3\textwidth]{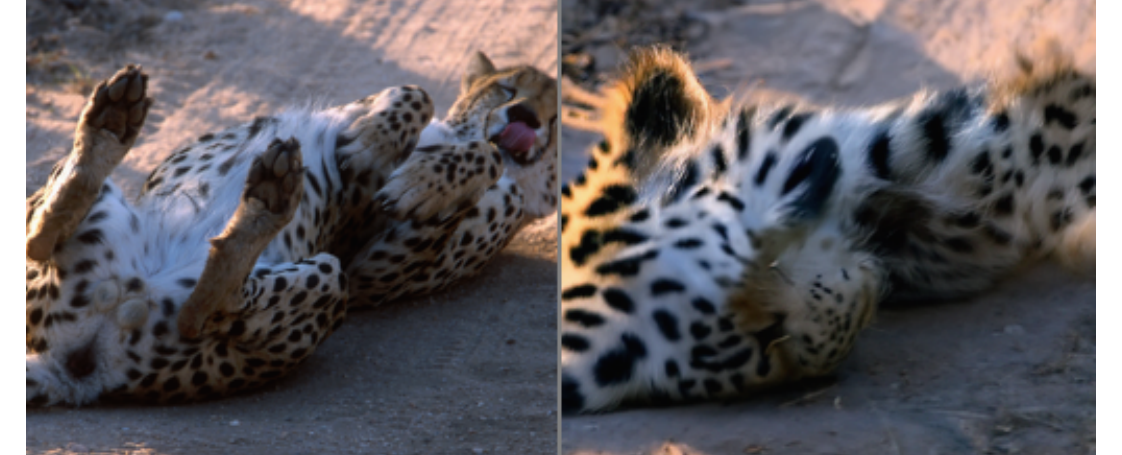} & \includegraphics[trim={1cm 11cm 1cm 11cm},clip,width=0.3\textwidth]{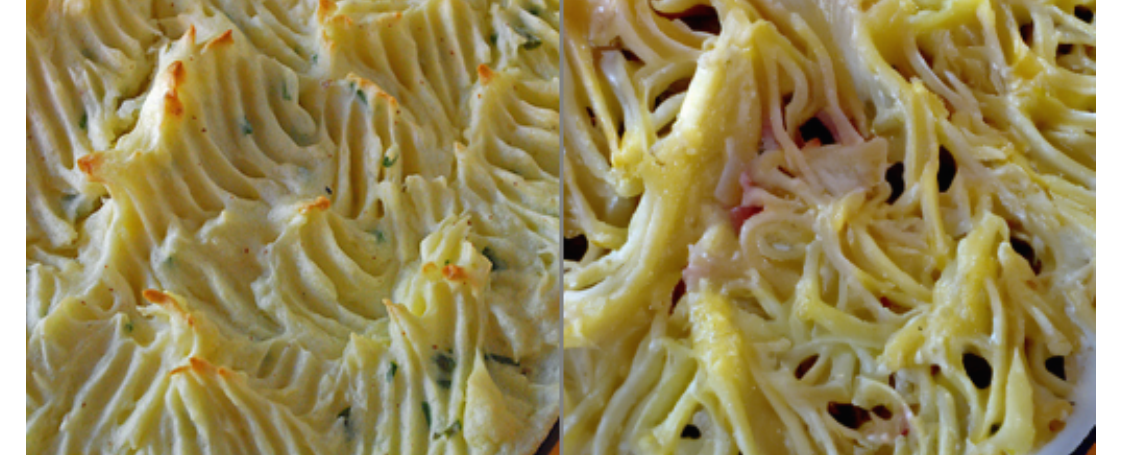} & \includegraphics[trim={1cm 11cm 1cm 11cm},clip,width=0.3\textwidth]{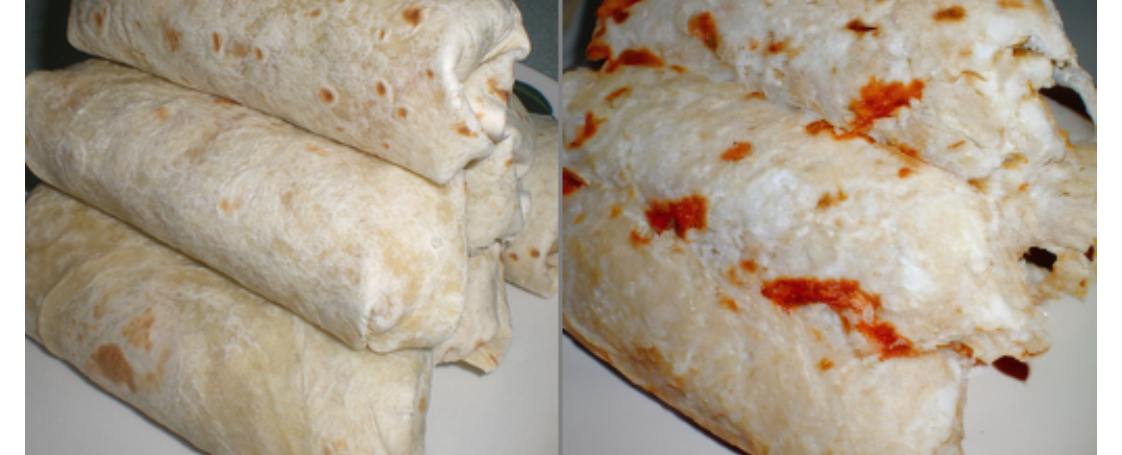} \\
    \shortstack{Generation\\Failed\\\vspace{0.3cm}} & \includegraphics[trim={1cm 11cm 1cm 11cm},clip,width=0.3\textwidth]{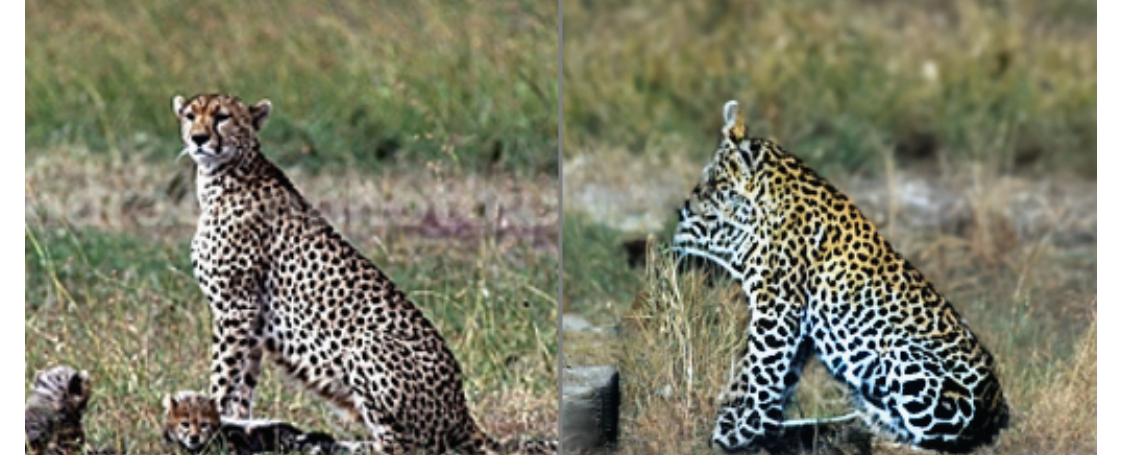} & \includegraphics[trim={1cm 11cm 1cm 11cm},clip,width=0.3\textwidth]{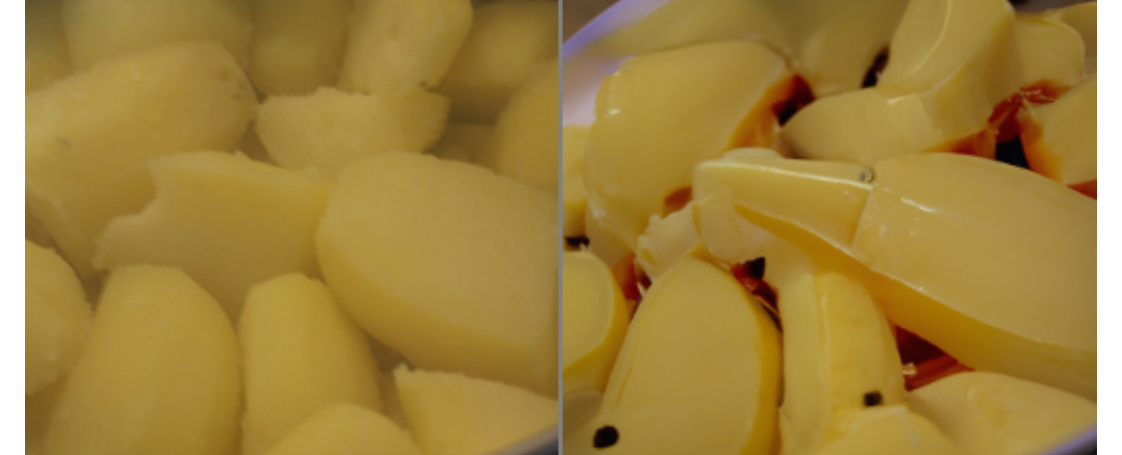} & \includegraphics[trim={1cm 11cm 1cm 11cm},clip,width=0.3\textwidth]{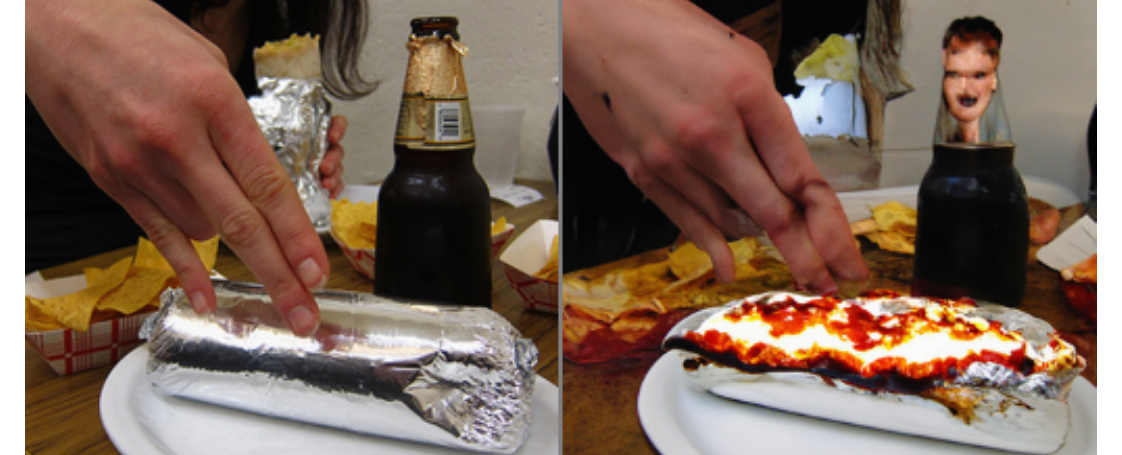} \\
    \bottomrule
    \end{tabular}
    \end{adjustbox}
    \medskip
    \caption{Hand-picked examples of pairs of original images and VCE attempts from our experiments.
    First row: desirable visual counterfactual explanations. 
    Second row: examples with changes not sufficient to flip to the target class. 
    Third row: method failures generating nonsensical outputs.
    } 
    \label{tab:vces}
\end{figure}

One of the difficulties of research in counterfactual methods is the lack of broadly accepted evaluation procedures.
The state-of-the-art are visual inspection of individual explanations and small-scale user studies.
While human expert judgments are crucial for model evaluation in domain context, they remain subjective and do not scale to typical large-scale image data sets.
Further, the showcased examples are subject to cherry-picking (such as row one in Table~\ref{tab:vces}). Instead, we recommend a comprehensive evaluation on all data, including failure cases (such as rows two and three in the same table).

In this paper we contribute to the discussion on best evaluation practices for VCE methods.
We review a number of previously proposed metrics and identify the most useful ones for assessing the properties of natural image VCEs (Section~\ref{sec:Metrics}).
Using these, we systematically quantitatively evaluate recent, diffusion-based, generative VCE methods \cite{dvce,dime} across a range of robust and non-robust classifiers (Section~\ref{sec:Evaluation}).
We complement this by an ablation of critical design choices that we identified as promising candidates for future improvements (Section~\ref{subsec:dvcedime}).
We provide access to our streamlined implementation of the VCE methods containing the ablation options as well as additional classifiers not present in the original code\footnote{\url{https://github.com/cairo-thws/DBVCE_eval}}.
In line with the best reproducibility practices, this codebase also includes all our scripts to execute the evaluation, replicate our results and recreate all the summary tables presented in this paper.
To foster transparency, we also provide access to our experimental tracking\footnote{\url{https://wandb.ai/fhws_cairo/DBVCE_eval}}.
Finally, we highlight the challenges related to setting up such an evaluation exercise on a conservative hardware configuration and, based on our experience, recommend an approach to tackle these (Section~\ref{sec:ComputeChallenges}).
We hope fellow researchers will benefit from these when setting up their own evaluation protocols.

\section{Visual Counterfactual Explanations through Diffusion Models}
In the following section we introduce the basic idea of diffusion models, specifically Denoising Diffusion Probabilistic Models (DDPM) \cite{ddpm}, explain the concept of classifier guidance, and draw the connection to counterfactual generation.
\subsection{Denoising Diffusion Probabilistic Models}\label{sec:DDPM}
The DDPM model is an encoder-decoder architecture, where the encoder is defined as a Markov chain linear Gaussian model $q(x_t | x_{t-1})$ progressively adding random Gaussian noise to an image $x_0$ with a pre-defined variance schedule $\beta_1, \ldots, \beta_T$ over the diffusion steps $x_1, ..., x_t, ..., x_T$ so that $x_T \sim \mathcal{N}(0, I)$.
The model learns to revert the diffusion process via the decoder $p_\theta\left(x_{t-1} \mid x_t\right)$, gradually removing noise from the signal $x_T$ in order to produce samples matching the original data $x_0$.
It can be shown \cite{understandingdiffusion}, that the reverse process is also Gaussian so that the model boils down to learning the mean and covariance of the reverse process 
$
    p_\theta\left(x_{t-1} \mid x_t\right)=\mathcal{N}\left(\mu_\theta\left(x_t, t\right), \Sigma_\theta\left(x_t, t\right)\right)
$.

\subsection{Classifier Guidance}\label{sec:ClasssifierGuidance}
The vanilla DDPM introduced in Section~\ref{sec:DDPM} enables synthesizing random samples from the learned underlying data distribution.
However, to generate class-specific data points, such as VCEs, the reverse transition has to be adapted.
Given a classifier $p_\phi(y \mid x_t)$, the mean of the denoising transition can be shifted by the gradients of the classifier w.r.t $x_t$ to yield class-conditional samples \cite{diffusionbeatgans}.
\begin{equation}\label{eq:ClassifierGuidance}
    \begin{aligned}
    p_{\theta, \phi}\left(x_{t-1} \mid x_t, y\right) & =\mathcal{N}\left(\mu_t, \Sigma_\theta\left(x_t, t\right)\right) \\
    \mu_t & =\mu_\theta\left(x_t, t\right) + s \cdot \Sigma_\theta\left(x_t, t\right) \nabla_{x_t} \log\big[p_\phi\left(y \mid x_t\right)\big] \enspace ,
    \end{aligned}
\end{equation}
where $s$ is a gradient scaling factor controlling the trade-off between sample diversity and class consistency.

\subsection{Diffusion-based Counterfactual Explanations}
\label{subsec:dvcedime}
We introduce two recent diffusion-based generative models for counterfactual explanations, DVCE \cite{dvce} and DiME \cite{dime}, focusing on the particular design choices related to adapting the diffusion models for XAI purposes.
The methods assume the availability of a DDPM pre-trained over the same images as the classifier $p_{\phi}$ to be explained.
Both of the methods propose to guide the learned decoder of the process via the classifier gradients similar to equation~\eqref{eq:ClassifierGuidance} to produce a realistic example classified by $p_{\phi}$ as the target counterfactual class.
However, to remain visually near the original image, both DVCE and DiME begin the denoising not from $x_T$, but from $x_{T/2}$, such that much of the rough structure of the original image is maintained but finer details can still be adapted to the target class.

One difficulty in using the classifier guidance of \cite{diffusionbeatgans} for VCEs of an arbitrary classifier $p_\phi$ is that $p_\phi$ cannot be expected to have been trained on noisy images.
Accordingly, $p_\phi$ cannot be expected to perform well on noisy data $x_t$ generated during the reverse diffusion process and provide sensible gradients $\nabla_{x_t} \log p_\phi\left(y \mid x_t\right)$ for the guidance.
To tackle this issue, the authors in both the DVCE and DiME paper propose to replace the noisy example $x_t$ in the input of the classifier by an approximation of the denoised image $\hat{x}_0 = x_\theta(x_t, t)$, where $x_\theta$ is the learned DDPM denoising function. 
We call this important deviation from the original formulation of the classifier guidance \cite{diffusionbeatgans} the \xzeropred approach and explore its effects more closely in Section~\ref{sec:Evaluation}.

While the \xzeropred approach solves the mismatch between the classifier and the noised diffusion data distributions, it presents an additional challenge for the gradient-based guidance of the sampling process.
In principle, it requires to back-propagate not only through the classifier $p_\phi$ but also through the diffusion process $p_\theta$ (from $\hat{x}_0$ to $x_t$) to obtain the gradients with respect to the noised $x_t$.
This greatly increases the computational as well as the memory requirements for generating VCEs. 
We discuss the related technical challenges and our approach to tackle them with a fairly conservative hardware configuration in Section~\ref{sec:ComputeChallenges}.

Another important design choice introduced in the DVCE paper is the \coneprojection approach.
The authors argue that gradients of non-robust classifiers are noisy, making the guided generation of semantically meaningful images particularly difficult.
Therefore, they propose to use the gradients of an independent, pre-trained, robust classifier $p_r$, which are expected to be less noisy, to help in the guidance.
This is achieved by replacing the gradients of the classifier to be explained $\nabla_{x_t} \log p_\phi$ in equation~\eqref{eq:ClassifierGuidance} with those of the robust classifier $\nabla_{x_t} \log p_r$, projected onto a cone of $30^{\circ}$ around the gradients $\nabla_{x_t} \log p_\phi$.
This reduces the noise of the non-robust classifier gradients and steers the counterfactual generation towards more semantically meaningful changes.
We discuss the effects of the \coneprojection in more detail in Section~\ref{sec:Evaluation}.

\section{Metrics}\label{sec:Metrics}

The desirable properties of counterfactual explanations can be broadly categorized into four axes: validity, closeness, realism and diversity \cite{dvce,fid_ex1,dime,cfeval}.
Though these are generally alluded to in the VCE literature, they are rarely systematically quantitatively evaluated. 
We describe these properties here below together with relevant metrics that can be used for their assessment.
We employ these metrics in Section~\ref{sec:Evaluation} to systematically and quantitatively evaluate sets of VCEs generated under varying conditions to document their usefulness and provide some novel insights into the functioning of the explored methods.

For the following, let $f(\cdot)$ be the classifier that shall be explained, $X$ the set of images of class $y$ for which we wish to create counterfactual explanations, $y_t$ the target class label, $c_f(\cdot)$ the counterfactual generation method trying to flip the label to $y_t$, and $\mathbbm{1}[\cdot]$ the indicator function.

\subsection{Validity}
A counterexample generated by $c_f(.)$ is considered a valid counterfactual if the classifier assigns it to the counterfactual target class $y_t$.

\begin{wrapfigure}{r}{0.45\textwidth}
\centering \includegraphics[trim={1.7cm 3cm 1.7cm 2.5cm},clip,width=0.45\textwidth]{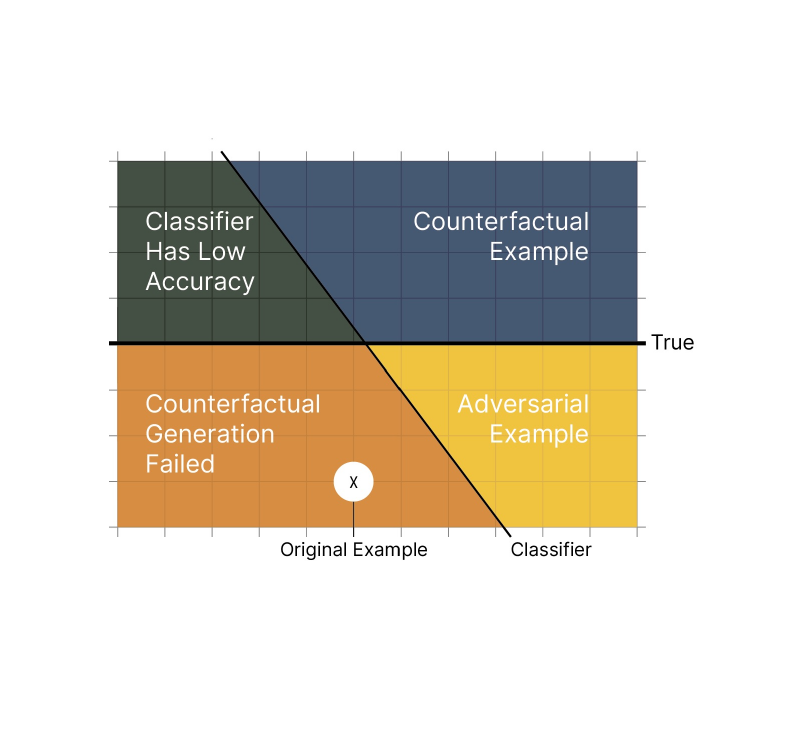}
\caption{Classifier class boundary versus true (oracle) class boundary and the interpretation thereof.}
\label{fig:cf_adv}
\end{wrapfigure}
A concept closely related to counterfactuals is that of adversarial examples \cite{goodfellow_adv}.
These are both perturbations of the original data example in such a way, that the classifier changes its decision.
There is, however, a significant difference between the two.
While the goal of an adversarial attack is to fool the classifier without actually changing the class of the example, a counterfactual shall truly be a realistic example of the target class.
In this paper we use the \emph{misclassification} criterion \cite{freiesleben_intriguing_2022} as the distinguishing property between the two concepts and illustrate the difference in Figure~\ref{fig:cf_adv}.

To explore the validity of the generated counterexamples, we propose to use a set of metrics based on the classification accuracy.

\emph{Target Accuracy}, $\mathrm{TA} = \frac{1}{|X|} \sum_{x \in X} \mathbbm{1}[f(\,c_f(x)\,)\, =\, y_t]$, is the fraction of generated counterexamples classified by $f$ to be within the target class $y_t$.

\emph{Original Accuracy}, $\mathrm{OA} = \frac{1}{|X|} \sum_{x \in X} \mathbbm{1}[f(\,c_f(x)\,)\, =\, y]$, is the fraction of generated counterexamples classified by $f$ to be within the original class $y$.
For these examples, the image changes are not sufficient to flip the classification result.

For a multi-class data set, the generator may produce examples that the classifier attributes to neither the original $y$ nor the target class $y_t$.
The percentage of these is clearly $1-\mathrm{TA}-\mathrm{OA}$.

To confirm that the generated counterexamples classified by $f$ as $y_t$ are really counterfactuals rather than just adversarial examples, we employ a committee of alternative classification models $f_o$, trained on the same data set, to serve as an oracle replacing human-expert assessment.

\emph{Oracle score}, $\mathrm{OS}=\frac{1}{|X|} \sum_{x \in X} \mathbbm{1}[f(\,c_f(x)\,)\, =\, f_o(\,c_f(x)\,)]$, is the percentage of generated examples that are classified the same by both the classifier $f$ and the oracle $f_o$ \cite{cfeval}, as a cheap though not perfect proxy for human validity assessment.

\emph{Oracle Target Accuracy}, $\mathrm{OTA} = \frac{1}{|X|} \sum_{x \in X} \mathbbm{1}[f_o(\,c_f(x)\,)\, =\, y_t]$, is the percentage of counterexamples that are classified by the oracle $f_o$ to be within the target class $y_t$.
This might include examples for which the classifier $f$ does not flip the class to $y_t$ but still show some changes that convince the oracle $f_o$.

\subsection{Closeness}
For a generated counterexample to be considered a counterfactual, it shall be as close as possible to the original example.
Various forms of Minkowski metric $\mathrm{D_p} = \frac{1}{|X|} \sum_{x \in X} \lVert x - c_f(x)\rVert_p$ (e.g. $p \in {1, 1.5, 2}$) can be utilized to measure the distance. 

A metric corresponding better to human visual perception is the Learned Perceptual Image Patch Similarity (LPIPS) \cite{lpips}.
Instead of comparing the image examples in the original pixel space, it compares activations of image patches in a pre-trained network.
In our study we use the PyTorch implementation of TorchMetrics\footnote{\url{https://torchmetrics.readthedocs.io}}.

\subsection{Realism}
VCE shall not just contain class-discriminative artifacts flipping the predicted class, but they shall be realistic examples of the target class.
We propose to use the state-of-the-art metric for image generation quality, the Fréchet Inception Distance (FID) \cite{fid}, suggested in some previous XAI studies \cite{fid_ex1,fid_ex2,dvce,dime}.
The FID compares the distribution of the Inception V3 model \cite{inceptionv3} image embeddings between a set of original $x \in X$ images and their generated counterparts $\hat{x} = c_f(x)$.
We use the TorchMetrics implementation with 64 features.

\subsection{Diversity}\label{sec:diversity}
A property often disregarded by VCE methods \cite{fid_ex1} is the ability to generate multiple valid counterfactuals for a single analyzed image.
Differences within the set of these examples can provide richer information to the user and improve the understanding of the results.
To measure diversity, we propose to use pairwise LPIPS \cite{cfeval,lpips} between the set of generated VCEs.
Since the diffusion-based methods we analyze in this paper do not natively generate sets of counterfactuals, there is no efficient way to produce these.
Rerunning the experiments multiple times with different random seeds should produce some variations in the data and it would be interesting to evaluate their diversity.
Such large-scale experimentation is, however, beyond the capacity of our limited hardware setup and we leave it for future work.

\section{Experimental Setup}\label{sec:Experiments}

In our experiments, we systematically and quantitatively evaluate diffusion-generated VCEs (see Section~\ref{subsec:dvcedime}) across a suite of classifiers.
We analyze similar data as those presented in the DVCE paper but, rather than manually selecting a handful of examples for visual inspection, we calculate the relevant metrics (Section~\ref{sec:Metrics}) across a representative set of generations.

The generative diffusion model has been pre-trained on the unlabeled ImageNet (2012) data set \cite{ILSVRC15}.
Same as the DVCE paper we use the OpenAI implementation of guided diffusion\footnote{\url{https://github.com/openai/guided-diffusion}}.
We choose six classifiers pre-trained on the labeled version of the data set summarized in Table~\ref{tab:classifiers} complemented by a RandomNet, which is an AlexNet architecture with randomly initialized weights.
\begin{table}[]
\centering
\begin{adjustbox}{max width=\textwidth}
\begin{tabular}{lccccccc}
\toprule
classifier &        Madry &      AlexNet &        UnetE &     ConvNeXt &      SwinTFL &       SIMCLR &  RandomNet \\
\midrule
Name & MNR-RN50 & AlexNet & UNet & ConvNeXt-L & Swin-L &  SIMCLR & AlexNet \\ 
Parameters & $25$M & $62$M & $54$M & $350$M & $197$M & $94$M & $62$M \\
Accuracy & 57.9\% & 63.3\% & NA & 87.8\% & 87.3\% & 74.2\% 	& 0\% \\
Robustness &  \cmark &      \xmark &        (\cmark) &     \xmark &      \xmark &       \xmark & \xmark \\
Used in DVCE & \cmark &      \xmark &       \xmark&         \cmark&      \cmark &       \xmark & \xmark \\
Published  &  \cite{madry} &  \cite{alexnet}  &  \cite{diffusionbeatgans} &  \cite{convnext} & \cite{swintf} &  \cite{simclr}  & -\\
\bottomrule
\end{tabular}
\end{adjustbox}
\medskip
\caption{Classifiers included in our study. NA: Not reported.}
\label{tab:classifiers}
\end{table}

In our analysis, we focus on the six source and eleven target classes presented in the DVCE paper:
source - \textit{cheetah, chesapeake bay retriever, mashed potato, coral reef, alp, burrito},
target - \textit{leopard, tiger, golden retriever, labrador retriever, guacamole, carbonara, alp, cliff, volcano, pizza, burrito}.
For each of the source classes we generate counterfactuals for all 50 original images from the validation data set (6 x 50 = 300 source images).
For each of the 50 source images per source class, we generate a counterfactual towards each of the 11 target classes.
We thus obtain a total of 3300 (= 6 x 50 x 11) synthesized examples split into 66 (= 6 x 11) source-target groups.

For each of the source classes there is a class within the target set identified by the DVCE paper as the most relevant for counterfactuals.
For example, for cheetah it is the leopard, for mashed potato it is the carbonara.
These have been identified based on the WordNet similarity of the class labels.
In our view, these choices of ``ideal'' source-target pairs are rather arbitrary and other source-target pairs may be required in practice.
Accordingly, while most of our evaluation in Section~\ref{sec:Evaluation} is performed over these ``ideal'' pairs, we also explore the performance when other classes are chosen for the counterfactual generations.

We explore the effects of two critical design choices explained in Section~\ref{subsec:dvcedime}.
Firstly, we let the gradient guidance be based on classifying the point $x_t$ either with the \xzeropred (using $\hat{x}_0(x_t, t)$) or without (as in equation~\eqref{eq:ClassifierGuidance}).
Secondly, we use the \coneprojection of the robust classifier gradients or we drop it and rely solely on the gradient directions of the classifier that shall be explained.
To cover these four architectural combinations, we need to repeat all our experiments four times leading to a total of 13'200 generated counterexamples, organized into the 66 source-target groups (split by the ablation choices).
All these experiments have to be repeated for each classifier in our study, totaling 92'400~(=~13'200~x~7) diffusion-based generations.

\subsection{Challenges}\label{sec:ComputeChallenges}
All experiments are conducted on a single compute node with four Nvidia A100 GPUs with 80 GB of video memory each. 
We re-implemented the \xzeropred and the \coneprojection to fit our evaluation procedure and make multiple adjustments (e.g. CPU offloading, deleting gradients, clearing CUDA cache) to streamline the code and its execution (see the released code base).

A major challenge for our experiments are the GPU/VRAM requirements.
The classifiers as well as the diffusion model are large neural networks with millions of parameters\footnote{According to \cite{ddpm}, the DDPM model has 554M parameters.} (Table~\ref{tab:classifiers}) requiring several gigabytes of VRAM.
The highest memory trace is, however, due to the gradient calculation and especially that of the diffusion model due to the \xzeropred approach.
This is further exacerbated by the \coneprojection which needs to use two classifiers at a time.

We rely on the DataParallel functionality of PyTorch and spread the generation of the counterfactuals for a single batch of 50 examples (from a single source class) across all four GPUs.
Very large models such as UnetE and ConvNeXt had to be placed onto a dedicated GPU while the remaining three GPUs were used in parallel for the diffusion model.
This setup required further adaptations in the code (available within our released code base).

Given the high memory requirements, further parallelization was not possible and we had to run the 28 experiments sequentially (4 ablation options x 7 classifiers).
The total runtime for these was approximately two days (wall-clock time).
Some evaluation metrics could be calculated right after the generation step as averages across the batch of generated images.
Many more (oracles, LPIPS, FID) had to be calculated in a post-processing step as they required loading additional models into memory.

\section{Experimental Results} \label{sec:Evaluation}

In this section, we evaluate the generated VCEs using the metrics defined in Section~\ref{sec:Metrics}.
We calculate the metrics for each source-target group (indicated in Section~\ref{sec:Metrics} as the set $X$) and, for the final presentation, calculate averages and standard deviations across the seven source classes.

\subsection{Minimal Set of Metrics} \label{sec:MinimalMetrics}

In Table~\ref{tab:dvce_allmetrics} we report the complete set of metrics in the base setup of the experiments: ``ideal'' source-target pairs (as defined by \cite{dvce}), with \xzeropred, and with \coneprojection\!.
We first review these metrics and reduce the list to a recommended minimal set for further evaluation.

\begin{table}[]
    \setlength{\tabcolsep}{6pt}
    \centering
    \begin{adjustbox}{max width=\textwidth}
        \begin{tabular}{lcccccccc}
\toprule
classifier &        Madry &      AlexNet &        UnetE &     ConvNeXt &      SwinTFL &       SIMCLR &    RandomNet \\
metric          &              &              &              &              &              &              &              \\
\midrule
TA ↑            &   0.84 ± 0.1 &  0.99 ± 0.01 &  0.99 ± 0.01 &  0.95 ± 0.04 &   0.9 ± 0.08 &  0.45 ± 0.29 &    0.0 ± 0.0 \\
OA ↓            &  0.07 ± 0.07 &    0.0 ± 0.0 &  0.01 ± 0.01 &  0.02 ± 0.02 &  0.05 ± 0.03 &    0.0 ± 0.0 &  0.01 ± 0.01 \\
OS Madry ↑      &            - &  0.32 ± 0.14 &  0.36 ± 0.23 &  0.29 ± 0.17 &  0.32 ± 0.15 &  0.05 ± 0.03 &    0.0 ± 0.0 \\
OTA Madry ↑     &            - &  0.32 ± 0.14 &  0.36 ± 0.23 &  0.28 ± 0.16 &   0.3 ± 0.16 &   0.2 ± 0.13 &  0.21 ± 0.16 \\
OS AlexNet ↑    &  0.72 ± 0.13 &            - &  0.48 ± 0.21 &  0.36 ± 0.13 &  0.42 ± 0.13 &  0.02 ± 0.03 &   0.0 ± 0.01 \\
OTA AlexNet ↑   &  0.72 ± 0.14 &            - &  0.48 ± 0.21 &  0.36 ± 0.13 &  0.41 ± 0.15 &  0.15 ± 0.05 &  0.23 ± 0.11 \\
OS RandomNet ↓  &    0.0 ± 0.0 &    0.0 ± 0.0 &    0.0 ± 0.0 &    0.0 ± 0.0 &    0.0 ± 0.0 &    0.0 ± 0.0 &            - \\
OTA RandomNet ↓ &    0.0 ± 0.0 &    0.0 ± 0.0 &    0.0 ± 0.0 &    0.0 ± 0.0 &    0.0 ± 0.0 &    0.0 ± 0.0 &            - \\
OS UnetE ↑      &  0.72 ± 0.08 &  0.43 ± 0.15 &            - &  0.43 ± 0.19 &  0.48 ± 0.19 &  0.06 ± 0.04 &   0.0 ± 0.01 \\
OTA UnetE ↑     &  0.76 ± 0.07 &  0.43 ± 0.15 &            - &   0.4 ± 0.19 &   0.45 ± 0.2 &  0.17 ± 0.09 &  0.22 ± 0.13 \\
OS ConvNeXt ↑   &   0.72 ± 0.1 &  0.38 ± 0.14 &  0.57 ± 0.22 &            - &  0.69 ± 0.13 &  0.03 ± 0.04 &    0.0 ± 0.0 \\
OTA ConvNeXt ↑  &  0.69 ± 0.13 &  0.38 ± 0.14 &  0.57 ± 0.22 &            - &  0.64 ± 0.17 &  0.08 ± 0.05 &  0.17 ± 0.12 \\
OS SwinTFL ↑    &   0.7 ± 0.08 &  0.39 ± 0.23 &  0.53 ± 0.23 &  0.56 ± 0.22 &            - &  0.01 ± 0.02 &    0.0 ± 0.0 \\
OTA SwinTFL ↑   &  0.66 ± 0.12 &  0.39 ± 0.23 &  0.53 ± 0.23 &  0.54 ± 0.23 &            - &  0.07 ± 0.06 &  0.14 ± 0.13 \\
$\mathrm{D_1}$ (m) ↓        &  1.49 ± 0.21 &  1.23 ± 0.22 &  1.28 ± 0.21 &   1.2 ± 0.22 &  1.22 ± 0.22 &  1.16 ± 0.24 &  1.14 ± 0.23 \\
$\mathrm{D_{1.5}}$ (k) ↓      &  9.08 ± 1.16 &  7.55 ± 1.22 &  7.81 ± 1.12 &  7.31 ± 1.24 &  7.43 ± 1.24 &  7.08 ± 1.33 &  6.97 ± 1.27 \\
$\mathrm{D_2}$ (k) ↓        &  0.77 ± 0.09 &   0.64 ± 0.1 &  0.66 ± 0.08 &   0.62 ± 0.1 &   0.63 ± 0.1 &    0.6 ± 0.1 &   0.59 ± 0.1 \\
LPIPS ↓         &  0.38 ± 0.04 &  0.32 ± 0.05 &  0.32 ± 0.03 &  0.29 ± 0.04 &  0.29 ± 0.05 &  0.34 ± 0.07 &  0.26 ± 0.03 \\
FID ↓           &  1.87 ± 0.77 &  1.16 ± 0.83 &  0.93 ± 0.45 &    0.8 ± 0.4 &  0.94 ± 0.39 &  3.16 ± 2.95 &  1.18 ± 0.52 \\
\bottomrule
\end{tabular}
    \end{adjustbox}
    \medskip
    \caption{Comprehensive quantitative evaluation of the VCEs in the base setup.}
    \label{tab:dvce_allmetrics}
\end{table}

The critical metric for counterfactual validity is the TA -- a generated example is not a counterfactual if it has not flipped the classifier decision.
OA can provide additional information for failure cases but we deem it more useful for early development stages of novel methods since reasonably well-tuned counterfactual methods shall be able to move out of the original class.
We therefore propose to drop OA from the minimal list. 

We argued in Section~\ref{sec:Metrics} for the use of oracle classifiers $f_o$ to discriminate counterfactuals from adversarial attacks.
Since each single classifier may be prone to error, we further extend the concept into a set of classifiers as a committee of oracles.
A possible alternative would be merging all these into a single oracle in an ensemble-like manner.
The previously suggested OS \cite{cfeval} is in our view not fit for purpose as it measures agreements and disagreements irrespective of the class.
We therefore recommend the simpler OTA which focuses directly on the target class.

The Minkowski distance metrics $D_p$ are, on occasion, used as regularization terms for both counterfactual and adversarial generations.
Such per-pixel distances, however, do not well align with human perception \cite{lpips} and are therefore of little use for measuring the quality of \emph{visual} counterfactual explanations.
We propose to exclude them from the shortlist and use the LPIPS \cite{lpips} instead.

FID \cite{fid} is the only metric in our list for measuring realism, hence we recommend its inclusion.

Overall, we recommend the TA and OTA to evaluate validity, the LPIPS for closeness, and the FID measure for realism of VCEs. 
Accordingly, we report these metrics in the remainder of this section.

\subsection{Evaluation of Diffusion-based VCEs} \label{sec:EvaluationDBVCE}

We structure our analysis around five hypotheses about the functioning of the method.

\begin{table}[]
    \setlength{\tabcolsep}{4pt}
    \centering
    \begin{adjustbox}{max width=\textwidth}
        \begin{tabular}{lcccccccc}
\toprule
classifier &                 Madry &               AlexNet &                 UnetE &            ConvNeXt &      SwinTFL &       SIMCLR &             RandomNet \\
metric         &                       &                       &                       &                     &              &              &                       \\
\midrule
TA ↑           &            0.84 ± 0.1 &  \textbf{0.99 ± 0.01} &           0.99 ± 0.01 &         0.95 ± 0.04 &   0.9 ± 0.08 &  0.45 ± 0.29 &             0.0 ± 0.0 \\
OTA Madry ↑    &                     - &           0.32 ± 0.14 &  \textbf{0.36 ± 0.23} &         0.28 ± 0.16 &   0.3 ± 0.16 &   0.2 ± 0.13 &           0.21 ± 0.16 \\
OTA AlexNet ↑  &  \textbf{0.72 ± 0.14} &                     - &           0.48 ± 0.21 &         0.36 ± 0.13 &  0.41 ± 0.15 &  0.15 ± 0.05 &           0.23 ± 0.11 \\
OTA UnetE ↑    &  \textbf{0.76 ± 0.07} &           0.43 ± 0.15 &                     - &          0.4 ± 0.19 &   0.45 ± 0.2 &  0.17 ± 0.09 &           0.22 ± 0.13 \\
OTA ConvNeXt ↑ &  \textbf{0.69 ± 0.13} &           0.38 ± 0.14 &           0.57 ± 0.22 &                   - &  0.64 ± 0.17 &  0.08 ± 0.05 &           0.17 ± 0.12 \\
OTA SwinTFL ↑  &  \textbf{0.66 ± 0.12} &           0.39 ± 0.23 &           0.53 ± 0.23 &         0.54 ± 0.23 &            - &  0.07 ± 0.06 &           0.14 ± 0.13 \\
LPIPS ↓        &           0.38 ± 0.04 &           0.32 ± 0.05 &           0.32 ± 0.03 &         0.29 ± 0.04 &  0.29 ± 0.05 &  0.34 ± 0.07 &  \textbf{0.26 ± 0.03} \\
FID ↓          &           1.87 ± 0.77 &           1.16 ± 0.83 &           0.93 ± 0.45 &  \textbf{0.8 ± 0.4} &  0.94 ± 0.39 &  3.16 ± 2.95 &           1.18 ± 0.52 \\
\bottomrule
\end{tabular}

    \end{adjustbox}
    \medskip
    \caption{Quantitative evaluation of the VCEs in the base setup.}
    \label{tab:dvce_results}
\end{table}

\req{Hypothesis 1: The diffusion-based counterfactual method in its base setup (for ``ideal'' source-target pairs, with \coneprojection and \xzeropred) produces valid, close and realistic VCEs even for non-robust classifiers of diverse model capacity.} 
Table~\ref{tab:dvce_results} summarizes the performance for the seven classification algorithms in our study.
As documented by high TA, most classifiers achieve high proportions of correctly classified counterexamples.
Zero validity of RandomNet is in line with the inability of this untrained network to classify the images correctly.
Somewhat surprisingly, the validity of SIMCLR counterfactuals is also fairly low, around 45\% with very high standard deviation between the 6 original classes.
This suggests that, for some of the original classes, SIMCLR manages to generate high proportions of counterexamples within the target class while for others it dramatically fails.
We speculate that this failure is related to the self-supervised nature of the SIMCLR.
Unlike the other models, its hidden features are not attuned to the specific classification task that is being explained by the counterfactual method.
The effects of this difference in the feature learning on the counterfactual generation, however, require more detailed investigation.

On the other hand, the oracle accuracy OTA for most classifiers and oracles is well below 50\%.
We can conclude that the generative process is successfully guided by the classifier to be explained but, rather than valid counterfactuals, produces adversarial examples which are not of the expected target class (see examples in Table~\ref{tab:vces}).
The exception here is Madry, the only robust classifier in our set, which achieves around 70\% oracle accuracy, actually flipping the images to the target class.

While these conclusions contradict our Hypotheses 1, the low OTA scores shall not be interpreted unfavorably for the counterfactual method. 
The method shall help the users establish trust in the classifiers by explaining their behavior.
From these metrics, we can see that most classifiers in our set rely on features that are not truly consequential for the class. 
By modifying these within the images, the classifiers flip their decisions but fail to truly shift the image conceptually to the target class.
This shall lead into reducing the trust of the user in the classifier results (despite possibly high classification accuracy).

The closeness measured by LPIPS ranges between 0.29 to 0.32 for all classifiers except Madry, which leads to rather higher distance (0.38) between the origin-target pairs on average.
This indicates that smaller perceptual changes may be sufficient to fool the classifier under study, but not sufficient to genuinely flip the class (which only Madry achieves).

We use the FID scores to measure the realism of the counterexamples.
It is the ConvNeXt guidance that generates examples which are the most coherent with the underlying probability distribution of the target class.
Since this method also achieves rather small LPIPS, it leads us to belief that only meaningful changes get introduced into the images but these are either too small or not semantically coherent to flip the decision of the oracles.
Madry, in contrast, suffers from much higher mismatch between the distributions of the true and generated images.
This backs the LPIPS conclusions: Madry guidance introduces more important changes into the images sometimes pushing them further away from the underlying data distribution.
SIMCLR has by far the largest FID as it introduces changes that are often not coherent with the target class at all (low TA and OTA).
This reconfirms the failure of the counterfactual method on this self-supervised classifier.

\req{
    Hypothesis 2: No meaningful changes are introduced into the images when generating counterexamples for a random classifier.
}
The RandomNet results in Table~\ref{tab:dvce_results} confirm that, as expected, the method produces no valid counterfactuals (zero TA).
Nevertheless, some small changes are introduced (low LPIPS) keeping the images realistic (low FID) but, surprisingly, switching the class of almost 20\% of the images (OTA scores).
This contradicts our Hypotheses 2.
Given that the classifier is random and its gradients toward the target class are therefore meaningless, this is a rather undesirable behavior which can be explained by the robust classifier feeding the method through the \coneprojection with semantically meaningful gradient directions.
This could, in practice, be resolved by introducing a rejection step into the counterfactual method, dropping all generated examples that are not re-classified by the classifier to be explained into the target class.
We recommend to adopt this step as a simple fix into the VCE methods in the future.
Based on the above analysis, we advocate for systematic inclusion of similar edge-case experiments into the evaluation of XAI methods as these can lead to useful unexpected insights.

\begin{table}[]
    \setlength{\tabcolsep}{4pt}
    \centering
    \begin{adjustbox}{max width=\textwidth}
        \begin{tabular}{lcccccccc}
\toprule
classifier &                                            Madry &                                         AlexNet &                                           UnetE &                                         ConvNeXt &                                         SwinTFL &                                           SIMCLR &                                       RandomNet \\
metric         &                                                  &                                                 &                                                 &                                                  &                                                 &                                                  &                                                 \\
\midrule
TA ↑           &    \colorgrey 0.00 \colorblack ± \colorgrey 0.00 &  \colorred -0.01 \colorblack ± \colorgreen 0.02 &  \colorred -0.04 \colorblack ± \colorgreen 0.05 &   \colorred -0.12 \colorblack ± \colorgreen 0.06 &  \colorred -0.19 \colorblack ± \colorgreen 0.12 &   \colorgreen 0.12 \colorblack ± \colorred -0.03 &   \colorgrey 0.00 \colorblack ± \colorgrey 0.00 \\
OTA Madry ↑    &                                                - &   \colorred -0.29 \colorblack ± \colorred -0.11 &   \colorred -0.31 \colorblack ± \colorred -0.16 &    \colorred -0.27 \colorblack ± \colorred -0.14 &   \colorred -0.29 \colorblack ± \colorred -0.14 &    \colorred -0.19 \colorblack ± \colorred -0.11 &   \colorred -0.20 \colorblack ± \colorred -0.15 \\
OTA AlexNet ↑  &   \colorgrey 0.00 \colorblack ± \colorgreen 0.01 &                                               - &   \colorred -0.27 \colorblack ± \colorred -0.05 &    \colorred -0.30 \colorblack ± \colorred -0.07 &   \colorred -0.30 \colorblack ± \colorred -0.07 &    \colorred -0.12 \colorblack ± \colorred -0.02 &   \colorred -0.19 \colorblack ± \colorred -0.07 \\
OTA UnetE ↑    &   \colorgrey 0.00 \colorblack ± \colorgreen 0.01 &   \colorred -0.37 \colorblack ± \colorred -0.13 &                                               - &    \colorred -0.35 \colorblack ± \colorred -0.10 &   \colorred -0.39 \colorblack ± \colorred -0.14 &    \colorred -0.15 \colorblack ± \colorred -0.08 &   \colorred -0.21 \colorblack ± \colorred -0.11 \\
OTA ConvNeXt ↑ &    \colorgrey 0.00 \colorblack ± \colorred -0.01 &   \colorred -0.30 \colorblack ± \colorred -0.09 &   \colorred -0.27 \colorblack ± \colorred -0.07 &                                                - &   \colorred -0.36 \colorblack ± \colorred -0.04 &    \colorred -0.06 \colorblack ± \colorred -0.04 &   \colorred -0.16 \colorblack ± \colorred -0.12 \\
OTA SwinTFL ↑  &    \colorred -0.01 \colorblack ± \colorgrey 0.00 &   \colorred -0.29 \colorblack ± \colorred -0.16 &   \colorred -0.26 \colorblack ± \colorred -0.08 &    \colorred -0.34 \colorblack ± \colorred -0.12 &                                               - &    \colorred -0.06 \colorblack ± \colorred -0.05 &   \colorred -0.14 \colorblack ± \colorred -0.14 \\
LPIPS ↓        &    \colorgrey 0.00 \colorblack ± \colorgrey 0.00 &   \colorred -0.04 \colorblack ± \colorgrey 0.00 &   \colorred -0.02 \colorblack ± \colorgrey 0.00 &    \colorred -0.04 \colorblack ± \colorgrey 0.00 &   \colorred -0.04 \colorblack ± \colorred -0.01 &   \colorred -0.01 \colorblack ± \colorgreen 0.01 &   \colorred -0.04 \colorblack ± \colorgrey 0.00 \\
FID ↓          &  \colorgreen 0.01 \colorblack ± \colorgreen 0.01 &   \colorred -0.12 \colorblack ± \colorred -0.04 &  \colorgreen 0.42 \colorblack ± \colorred -0.04 &  \colorgreen 0.17 \colorblack ± \colorgreen 0.34 &  \colorred -0.07 \colorblack ± \colorgreen 0.37 &  \colorgreen 1.18 \colorblack ± \colorgreen 1.66 &  \colorgreen 0.36 \colorblack ± \colorred -0.06 \\
\bottomrule
\end{tabular}

    \end{adjustbox}
    \medskip
    \caption{Change in metrics as compared to Table~\ref{tab:dvce_results}: no \coneprojection\!.}
    \label{tab:dvce_nocone_results}
\end{table}

\req{
    Hypothesis 3: Cone projection is the critical factor for generating valid counterfactuals rather than just adversarial examples.
}
In Table~\ref{tab:dvce_nocone_results}, we report the changes in the metrics (increases and decreases) as compared to Table~\ref{tab:dvce_results} when the guidance is based solely on the gradients of the classifier to be explained without any robust classifier \coneprojection\!.

The results for the robust classifier Madry remain unchanged in this scenario because \coneprojection merely projects Madry's gradients onto  its own cone (thus leaving them unchanged).
Similarly, the target accuracy of the random classifier also remains at zero.
However, the OTA decrease corroborates our conclusions from the previous section on the undesirable effects of the robust classifier \coneprojection for this edge-case.

For the other classifiers, the target accuracy drops (slightly for the AlexNet and UnetE, rather more for ConvNetXt and SwinTFL).
OTA drops significantly across the board meaning that though the counterexamples produced by the method without the \coneprojection fool the classifiers to be explained, they are not valid counterfactuals as they do not de facto belong to the target class (low OTA).
This confirms our Hypotheses 3 about the significance of the \coneprojection step in preventing adversarial examples.

Similar to prior analysis, the behavior of SIMCLR stands out. 
It is the only classifier for which the TA improves when not using \coneprojection\!.
However, given that OTA drops as well, this again signifies that mostly adversarial examples get generated.
From the increase in TA we surmise that the initial failure of the counterfactual method for SIMCLR reported in Table~\ref{tab:dvce_results} may be due to large disagreements between the SIMLCR and the robust classifier in the desirable gradient directions.
This will lead to frequent projections of the gradients on the edge of the cone finding a compromise direction which does not guide the process towards useful and meaningful changes. 

\begin{table}[]
    \setlength{\tabcolsep}{4pt}
    \centering
    \begin{adjustbox}{max width=\textwidth}
        \begin{tabular}{lcccccccc}
\toprule
classifier &                                           Madry &                                         AlexNet &                                           UnetE &                                         ConvNeXt &                                          SwinTFL &                                           SIMCLR &                                        RandomNet \\
metric         &                                                 &                                                 &                                                 &                                                  &                                                  &                                                  &                                                  \\
\midrule
TA ↑           &   \colorgrey 0.00 \colorblack ± \colorgrey 0.00 &  \colorred -0.08 \colorblack ± \colorgreen 0.07 &   \colorgrey 0.00 \colorblack ± \colorgrey 0.00 &   \colorred -0.58 \colorblack ± \colorgreen 0.15 &   \colorred -0.55 \colorblack ± \colorgreen 0.18 &    \colorred -0.45 \colorblack ± \colorred -0.31 &    \colorgrey 0.00 \colorblack ± \colorgrey 0.00 \\
OTA Madry ↑    &                                               - &  \colorred -0.06 \colorblack ± \colorgreen 0.01 &   \colorred -0.02 \colorblack ± \colorgrey 0.00 &    \colorred -0.06 \colorblack ± \colorgrey 0.00 &   \colorred -0.08 \colorblack ± \colorgreen 0.01 &  \colorgreen 0.02 \colorblack ± \colorgreen 0.03 &    \colorgrey 0.00 \colorblack ± \colorgrey 0.00 \\
OTA AlexNet ↑  &   \colorred -0.01 \colorblack ± \colorred -0.02 &                                               - &   \colorred -0.02 \colorblack ± \colorred -0.02 &    \colorred -0.07 \colorblack ± \colorgrey 0.00 &    \colorred -0.12 \colorblack ± \colorred -0.01 &  \colorgreen 0.09 \colorblack ± \colorgreen 0.06 &  \colorgreen 0.02 \colorblack ± \colorgreen 0.01 \\
OTA UnetE ↑    &  \colorgrey 0.00 \colorblack ± \colorgreen 0.02 &   \colorred -0.09 \colorblack ± \colorred -0.01 &                                               - &    \colorred -0.15 \colorblack ± \colorred -0.06 &    \colorred -0.19 \colorblack ± \colorred -0.07 &  \colorgreen 0.05 \colorblack ± \colorgreen 0.04 &    \colorgrey 0.00 \colorblack ± \colorgrey 0.00 \\
OTA ConvNeXt ↑ &   \colorred -0.02 \colorblack ± \colorgrey 0.00 &  \colorred -0.10 \colorblack ± \colorgreen 0.01 &   \colorred -0.01 \colorblack ± \colorgrey 0.00 &                                                - &    \colorred -0.38 \colorblack ± \colorred -0.01 &  \colorgreen 0.07 \colorblack ± \colorgreen 0.04 &    \colorred -0.03 \colorblack ± \colorred -0.02 \\
OTA SwinTFL ↑  &  \colorgreen 0.01 \colorblack ± \colorgrey 0.00 &   \colorred -0.10 \colorblack ± \colorred -0.04 &  \colorgreen 0.04 \colorblack ± \colorgrey 0.00 &    \colorred -0.30 \colorblack ± \colorred -0.06 &                                                - &  \colorgreen 0.07 \colorblack ± \colorgreen 0.08 &    \colorred -0.01 \colorblack ± \colorred -0.01 \\
LPIPS ↓        &   \colorgrey 0.00 \colorblack ± \colorgrey 0.00 &   \colorred -0.03 \colorblack ± \colorred -0.01 &   \colorgrey 0.00 \colorblack ± \colorgrey 0.00 &    \colorred -0.03 \colorblack ± \colorgrey 0.00 &    \colorred -0.03 \colorblack ± \colorred -0.01 &    \colorred -0.08 \colorblack ± \colorred -0.04 &    \colorgrey 0.00 \colorblack ± \colorgrey 0.00 \\
FID ↓          &   \colorred -0.13 \colorblack ± \colorred -0.03 &   \colorred -0.18 \colorblack ± \colorred -0.19 &   \colorred -0.03 \colorblack ± \colorgrey 0.00 &  \colorgreen 0.37 \colorblack ± \colorgreen 0.17 &  \colorgreen 0.27 \colorblack ± \colorgreen 0.21 &    \colorred -1.96 \colorblack ± \colorred -2.66 &    \colorred -0.03 \colorblack ± \colorred -0.01 \\
\bottomrule
\end{tabular}

    \end{adjustbox}
    \medskip
    \caption{Change in metrics as compared to Table~\ref{tab:dvce_results}: no \xzeropred\!.}
    \label{tab:dvce_noxzeropred_results}
\end{table}

\req{
    Hypothesis 4: The \xzeropred enables the counterfactual generation on non-robust classifiers.
}
We argued in Section~\ref{subsec:dvcedime} that the \xzeropred enables the counterfactual generation for classifiers not trained on noisy images and therefore not able to correctly classify the noisy samples of the intermediate diffusion generative process.

In Table~\ref{tab:dvce_noxzeropred_results} we see that the \xzeropred has a negligible effect for Madry and UnetE classifiers which were both trained on noisy images.
For the random classifier, the effect is also almost zero, which is as expected and verifies our previous conclusion that the image changes in the standard setting are in fact caused by the \coneprojection\!.
The three non-robust classifiers ConvNeXt, SwinTFL and SIMCLR not trained on noisy images loose substantial target accuracy and some oracle target accuracy.
This confirms our Hypotheses 4.
Interestingly, the TA and OTA scores of AlexNet decrease only slightly. 
We hypothesize that this is due to the low model capacity of AlexNet which forces it to learn only a low number of class-discriminative features.
This makes the model, perhaps somewhat surprisingly, robust to noised images.

\req{
    Hypothesis 5: Counterfactual generation for badly chosen target classes fails or produces adversarial attacks.
}
\begin{table}[]
    \setlength{\tabcolsep}{4pt}
    \centering
    \begin{adjustbox}{max width=\textwidth}
        \begin{tabular}{lcccccccc}
\toprule
classifier &                                            Madry &                                          AlexNet &                                            UnetE &                                         ConvNeXt &                                          SwinTFL &                                           SIMCLR &                                        RandomNet \\
metric         &                                                  &                                                  &                                                  &                                                  &                                                  &                                                  &                                                  \\
\midrule
TA ↑           &   \colorred -0.16 \colorblack ± \colorgreen 0.12 &   \colorred -0.03 \colorblack ± \colorgreen 0.03 &   \colorred -0.09 \colorblack ± \colorgreen 0.09 &   \colorred -0.23 \colorblack ± \colorgreen 0.12 &   \colorred -0.35 \colorblack ± \colorgreen 0.16 &   \colorgreen 0.08 \colorblack ± \colorred -0.06 &    \colorgrey 0.00 \colorblack ± \colorgrey 0.00 \\
OTA Madry ↑    &                                                - &   \colorred -0.14 \colorblack ± \colorgreen 0.05 &    \colorred -0.19 \colorblack ± \colorred -0.05 &    \colorred -0.16 \colorblack ± \colorgrey 0.00 &    \colorred -0.17 \colorblack ± \colorgrey 0.00 &    \colorred -0.12 \colorblack ± \colorred -0.01 &    \colorred -0.13 \colorblack ± \colorred -0.02 \\
OTA AlexNet ↑  &   \colorred -0.23 \colorblack ± \colorgreen 0.10 &                                                - &    \colorred -0.29 \colorblack ± \colorred -0.03 &   \colorred -0.23 \colorblack ± \colorgreen 0.04 &   \colorred -0.24 \colorblack ± \colorgreen 0.02 &   \colorred -0.11 \colorblack ± \colorgreen 0.02 &    \colorred -0.17 \colorblack ± \colorgrey 0.00 \\
OTA UnetE ↑    &   \colorred -0.22 \colorblack ± \colorgreen 0.16 &   \colorred -0.22 \colorblack ± \colorgreen 0.04 &                                                - &    \colorred -0.25 \colorblack ± \colorred -0.02 &    \colorred -0.28 \colorblack ± \colorred -0.03 &   \colorred -0.11 \colorblack ± \colorgreen 0.03 &   \colorred -0.15 \colorblack ± \colorgreen 0.01 \\
OTA ConvNeXt ↑ &   \colorred -0.24 \colorblack ± \colorgreen 0.09 &   \colorred -0.16 \colorblack ± \colorgreen 0.04 &    \colorred -0.35 \colorblack ± \colorred -0.05 &                                                - &   \colorred -0.34 \colorblack ± \colorgreen 0.03 &   \colorred -0.05 \colorblack ± \colorgreen 0.05 &   \colorred -0.12 \colorblack ± \colorgreen 0.01 \\
OTA SwinTFL ↑  &   \colorred -0.21 \colorblack ± \colorgreen 0.11 &    \colorred -0.18 \colorblack ± \colorred -0.05 &    \colorred -0.33 \colorblack ± \colorred -0.05 &    \colorred -0.31 \colorblack ± \colorred -0.01 &                                                - &   \colorred -0.04 \colorblack ± \colorgreen 0.03 &    \colorred -0.09 \colorblack ± \colorgrey 0.00 \\
LPIPS ↓        &   \colorgreen 0.05 \colorblack ± \colorgrey 0.00 &   \colorgreen 0.06 \colorblack ± \colorgrey 0.00 &   \colorgreen 0.03 \colorblack ± \colorgrey 0.00 &   \colorgreen 0.02 \colorblack ± \colorgrey 0.00 &   \colorgreen 0.04 \colorblack ± \colorred -0.01 &   \colorgreen 0.02 \colorblack ± \colorred -0.01 &   \colorgreen 0.02 \colorblack ± \colorgrey 0.00 \\
FID ↓          &  \colorgreen 0.46 \colorblack ± \colorgreen 0.68 &  \colorgreen 0.41 \colorblack ± \colorgreen 0.25 &  \colorgreen 0.61 \colorblack ± \colorgreen 0.45 &  \colorgreen 0.46 \colorblack ± \colorgreen 0.37 &  \colorgreen 0.56 \colorblack ± \colorgreen 0.67 &  \colorgreen 0.34 \colorblack ± \colorgreen 1.36 &  \colorgreen 0.44 \colorblack ± \colorgreen 0.33 \\
\bottomrule
\end{tabular}

    \end{adjustbox}
    \medskip
    \caption{Change in metrics as compared to Table~\ref{tab:dvce_results}: non-``ideal'' targets.}
    \label{tab:badcfs}
\end{table}

The evaluation in Table~\ref{tab:badcfs} is conducted over the 10 non-``ideal'' source-target pairs for each of the 6 original classes.
TA, but mainly OTA, decrease rather dramatically while the LPIPS and FID increase. 
This indicates that the method struggles to produce valid counterfactuals for these target classes.
This is in line with our intuition, which suggests that it is fundamentally very difficult (and at the same time not very useful) to produce counterfactuals belonging to a class which is semantically very distant from the original class (e.g. from the animal world to food or physical geographical phenomenons).

These results trigger the question to what extent VCEs for natural scene classification even make sense (e.g. multiple concepts may be present within single image, the combination of the minor concepts may be inconsistent with major concept of the target class, etc.). 
At the same time, they clearly document that some target classes are more suitable for counterfactual generation than others. 
The choice of these lies in the hands of the analyst and for now there is little guidance on how this shall be achieved (DVCE used the WordNet similarity).
In our view both these questions deserve further attention in future research.

\section{Conclusion}
In this work, we advocate for more rigorous, quantitative assessment of visual counterfactual explanations (VCEs) in order to foster consistency and transparency of evaluation procedures and thus advance the understanding of the strength and weaknesses of different VCE methods.
We propose a framework and a set of metrics to be used, and we conduct an extensive study exploring the effects of critical design choices of diffusion-based VCE methods across a suite of classifiers.
Our investigation of the results illustrates how such a systematic quantitative analysis can provide valuable insights into the functioning of the methods and suggest directions for future improvements.
We hope that by sharing the results of our study and our experience with tackling the computational challenges, together with the complete code to replicate our results, we contribute to betterment of experimental and evaluation protocols in future reports on VCE methods.

An important limitation of our work is the lack of evaluation of the diversity in the counterfactual explanations.
We briefly outlined a possible approach to such an evaluation in Section~\ref{sec:diversity}.
However, given that generating multiple diverse counterfactuals for the same original image would require re-running the VCE method multiple times in sequence, the overall execution time proved to be prohibitively long for our study.
Nevertheless, we are convinced that having access to a diverse set of counterfactuals should greatly improve the user understanding of the classifier's decisions (as compared to a single counterfactual or a set of very similar ones).
In our future work we will investigate a more efficient experimental setup for our hardware configuration that would allow generation and evaluation of diverse counterfactuals more rapidly.
We also believe that the ability to generate diverse sets of counterfactuals is an important aspect of the VCE methods that deserves concentrated effort in future research.

\printbibliography

\end{document}